\documentclass[conference]{IEEEtran}
\IEEEoverridecommandlockouts

\usepackage{cite}
\usepackage{amsmath,amssymb,amsfonts}
\usepackage{graphicx}
\usepackage{textcomp}
\usepackage{xcolor}

\usepackage{multirow}
\usepackage{booktabs}
\usepackage{stfloats}
\usepackage{algpseudocode}
\usepackage{algorithm}
\usepackage[hidelinks]{hyperref}

\def\BibTeX{{\rm B\kern-.05em{\sc i\kern-.025em b}\kern-.08em
    T\kern-.1667em\lower.7ex\hbox{E}\kern-.125emX}}

\newcommand{\figref}[1]{Fig.~\ref{#1}}
\newcommand{\tabref}[1]{Tab.~\ref{#1}}

\newcommand{\secref}[1]{Sec.~\ref{#1}}

\newcommand{\eg}{\textit{e}.\textit{g}.}

\renewcommand{\thefootnote}{\fnsymbol{footnote}}

\begin{document}

\title{Contrastive Multi-Modal Hypergraph Reasoning for 3D Crowd Mesh Recovery}
\author{Minghao Sun$^{1,2}$* \hspace{5mm} Chongyang Xu$^{3}$* \hspace{5mm} Yitao Xie$^{1}$ \hspace{5mm} Buzhen Huang$^{1}$$\dagger$ \hspace{5mm} Kun Li$^{1}$\\%
\\
$^1$Tianjin University \hspace{1mm}
$^2$Nanyang Technological University \hspace{1mm}
$^3$Sichuan University\\
\\
\vspace{-13mm}
}

\maketitle

\footnotetext[1]{Equal contribution.}
\footnotetext[2]{Corresponding author: Buzhen Huang (hbz@tju.edu.cn). 

\thanks{This work was supported by the Science Fund for Distinguished Young Scholars of Tianjin under Grant 22JCJQJC00040.}
}
\renewcommand{\thefootnote}{\arabic{footnote}}

\begin{abstract}
Multi-person 3D reconstruction is pivotal for real-world interaction analysis, yet remains challenging due to severe occlusions and depth ambiguity. Current approaches typically rely on single-modality inputs, which inherently lack geometric guidance. Furthermore, these methods often reconstruct subjects in isolation, neglecting the collective group context essential for resolving ambiguities in crowded scenes. To address these limitations, we propose Contrastive Multi-modal Hypergraph Reasoning to synergize semantic, geometric, and pose cues for crowd reconstruction. We first initialize robust node representations by combining RGB features, geometric priors, and occlusion-aware incomplete poses. Additionally, we introduce a pelvis depth indicator as a global spatial anchor, aligning visual features with a metric-scale-agnostic depth ordering. Subsequently, we construct a shared-topology hypergraph that moves beyond pairwise constraints to model higher-order crowd dynamics. To improve feature fusion, we design a hypergraph-based contrastive learning scheme that jointly enhances intra-modal discriminability and enforces cross-modal orthogonality. This mechanism enables the network to propagate global context effectively, allowing it to infer missing information even under severe occlusion. Extensive experiments on the Panoptic and GigaCrowd benchmarks confirm that our method achieves new state-of-the-art performance. Code and pre-trained models are available at \url{https://github.com/SunMH-try/CoMHR}.
\end{abstract}
\section{Introduction}
\label{sec:intro}

Multi-person 3D reconstruction is a cornerstone task in computer vision, serving for applications in virtual reality~\cite{Wen2025DyCrowdTD}, crowd analysis~\cite{Wen2023Crowd3DTH, Huang2023ReconstructingGO, Wen2025DyCrowdTD}, and human interaction~\cite{Huang2024CloselyIH, Huang2025ReconstructingCH}. While impressive progress has been made in reconstructing individuals, real-world scenarios predominantly involve multiple people interacting in crowded environments. Recovering accurate 3D meshes in such settings remains a formidable challenge due to severe inter-person occlusions~\cite{Huang2022ObjectOccludedHS}, complex spatial arrangements~\cite{Sun2021PuttingPI}, and the inherent depth ambiguity~\cite{Yang2024DepthAU}.

Current human mesh recovery methods predominantly rely on single-modality inputs, such as RGB images or 2D keypoints~\cite{Baradel2024MultiHMRMW, Stathopoulos2024ScoreGuidedDF}. While sufficient for isolated individuals, this reliance becomes a bottleneck in dense crowds, where severe depth ambiguities and frequent inter-person occlusions are common. Traditional one-stage methods attempt to handle crowded scenes by estimating all subjects jointly in a single pass~\cite{Sun2020MonocularOR, Sun2021PuttingPI, Baradel2024MultiHMRMW}, but often fail in large-scale crowds due to spatial conflicts and limited modeling of complex interactions. Hypergraph-based approaches, such as GroupRec~\cite{Huang2023ReconstructingGO}, explicitly model group interactions, yet their dependence on single-modality RGB inputs restricts high-order relational reasoning and geometric constraints, reducing reconstruction accuracy. Some recent works incorporate auxiliary modalities like LiDAR~\cite{Chen2022FUTR3DAU}, Radar~\cite{Chen2022ImmFusionRM}, or trajectory cues~\cite{Wen2025DyCrowdTD}, but typically use shallow fusion schemes that cannot fully exploit cross-modal information. Moreover, instance-centric designs reconstruct each individual independently, neglecting collective group context and high-order cross-subject priors necessary to resolve ambiguities in crowded scenes.

\begin{figure}[t]
    \centering
    \includegraphics[width=0.95\linewidth]{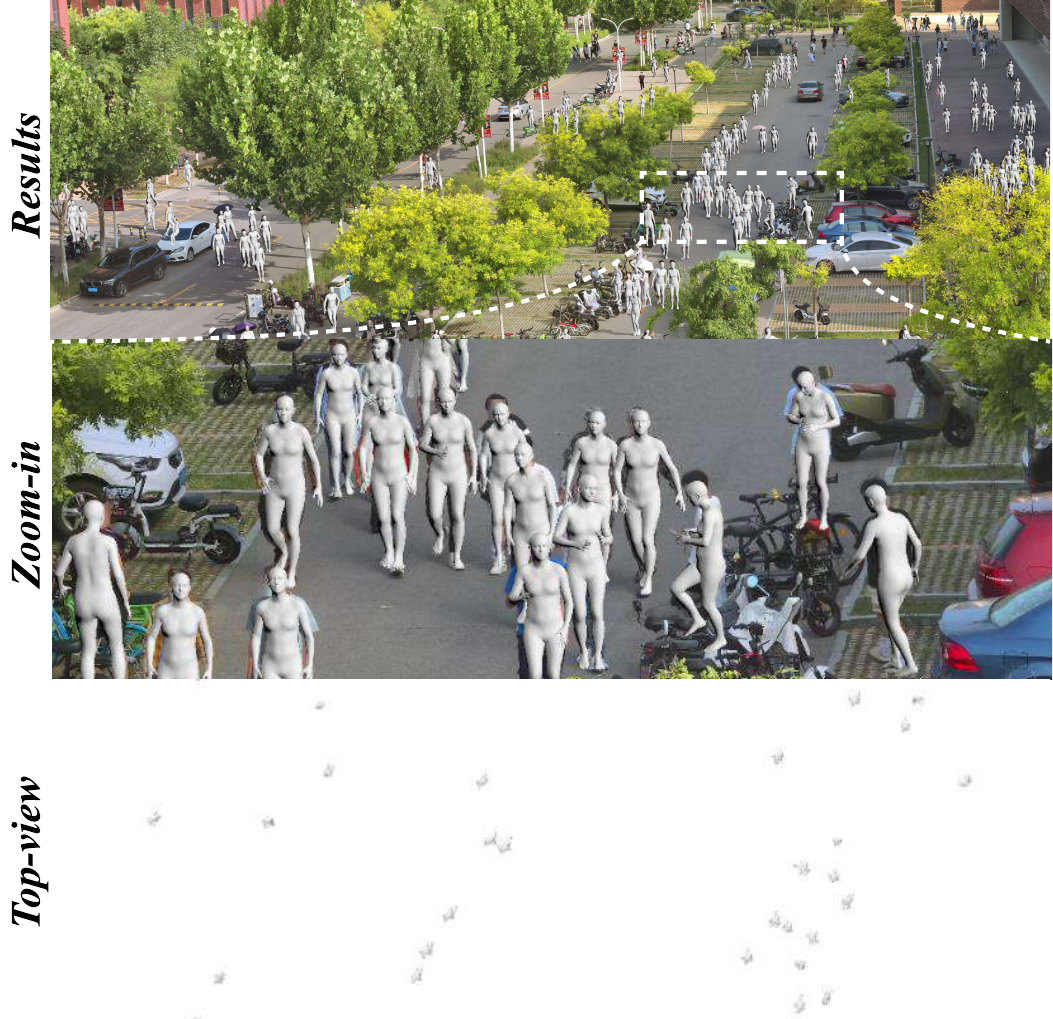}
    \vspace{-1.0em}
    \caption{Our method reconstructs spatially consistent human meshes in extremely dense crowd scenes, capturing accurate relative depth ordering.
    }
    \label{fig:teaser}
    \vspace{-1.0em}
\end{figure}

To address these gaps, we propose \textit{\textbf{Co}ntrastive \textbf{M}ulti-modal \textbf{H}ypergraph \textbf{R}easoning} (\textbf{CoMHR}), which exploits diverse multi-modal cues to enforce strong and robust constraints on individual poses and inter-person interactions. Our core insight is to synergize complementary modalities within a high-order topological structure, using contrastive learning to explicitly and effectively align and fuse heterogeneous features. To support this, we leverage robust geometric priors from foundation models. Although these models provide relative rather than metric depth, they serve as powerful pseudo-ground truth for constraining spatial relationships and depth ordering. By fusing robust geometric cues with RGB semantics and pose priors via a hypergraph~\cite{Feng2018HypergraphNN}, CoMHR captures high-order crowd interactions and relative depth ordering, ensuring robust reconstruction as illustrated in~\figref{fig:teaser}.

Specifically, we first initialize robust node representations by augmenting RGB features with relative geometric priors and occlusion-aware incomplete poses from foundation models~\cite{Yang2024DepthAU,Hidalgo2019OpenPoseW}. Crucially, to align these visual semantics with metric-scale-agnostic depth ordering, we also introduce \textit{Pelvis Depth Indicator} as a global spatial anchor, explicitly enforcing front-back ordering among individuals. Building upon these enriched representations, we construct a \textit{Shared-Topology Hypergraph} regularized by a \textit{Hypergraph Contrastive Learning} strategy designed to cluster individuals with similar actions \textit{(intra-modal)} and encourage orthogonality between heterogeneous modalities \textit{(cross-modal)}~\cite{Newell2017PixelsTG} to prevent feature contamination. With the feature space optimized, we execute high-order hypergraph reasoning to propagate global context across the shared topology, utilizing collective group dynamics to infer missing cues for occluded subjects. Finally, the refined representations facilitate accurate regression of pose and shape parameters, ensuring robustness under severe occlusion.
Our main contributions are summarized as follows:
\begin{itemize}
    \item We propose \textbf{Multimodal Hypergraph Reasoning} framework that synergizes multi-modal features to explicitly model high-order correlations for robust reconstruction.
    \item We introduce \textbf{Hypergraph Contrastive Learning} strategy to enhance intra-modal discriminability and cross-modal complementarity, ensuring feature robustness.
    \item Extensive experiments on public benchmarks demonstrate that our method achieves \textbf{state-of-the-art performance} with superior computational efficiency and robustness.
\end{itemize}
\section{Related Work}\label{sec:Related}

\subsection{Multi-Person Mesh Recovery}
Recent multi-person mesh recovery addresses depth ambiguity~\cite{Sun2021PuttingPI}, complex interactions~\cite{Huang2024CloselyIH, Huang2025ReconstructingCH}, and occlusion~\cite{Baradel2024MultiHMRMW}, typically via top-down~\cite{Moon2019CameraDT} or bottom-up~\cite{Sun2021PuttingPI, Baradel2024MultiHMRMW} strategies. However, most methods rely on single-modality 2D cues (\eg, RGB or keypoints) lacking explicit depth, causing fragility in crowds. While radar~\cite{Chen2022ImmFusionRM} and LiDAR~\cite{Chen2022FUTR3DAU} approaches exist, they require expensive hardware and often employ shallow fusion. Furthermore, high-order interaction modeling remains underexplored, as current group dynamics methods still depend on single modalities. To bridge these gaps, our multi-modal hypergraph framework unifies RGB, depth, and pose cues to model high-order dependencies, ensuring robust depth reasoning and mesh recovery in occluded scenes.

\subsection{Hypergraph Learning for Relational Modeling}
The HGNN framework~\cite{Feng2018HypergraphNN} introduced hyperedge convolution to encode high-order correlations, a method widely adopted in trajectory prediction~\cite{Xu2022GroupNetMH}, navigation~\cite{Li2024MultiAgentDR}, and pose estimation~\cite{Huang2023ReconstructingGO}. In 3D human reconstruction, hypergraphs have evolved from modeling non-local intra-person joint constraints~\cite{Hao2023HyperGraphBH} to capturing global crowd dynamics. Recent works~\cite{Huang2023ReconstructingGO} extend this to multi-person scenarios, using adaptive learning to represent interactions and occlusions. However, these methods remain limited by their reliance on single-modality features, leading to ambiguity in complex crowds. In contrast, our multi-modal contrastive framework leverages complementary cues to enhance robustness, securing resilient performance in highly occluded crowd scenes where single-modality approaches typically fail.

\begin{figure*}[htbp]  
    \centering
    \includegraphics[width=0.9\textwidth]{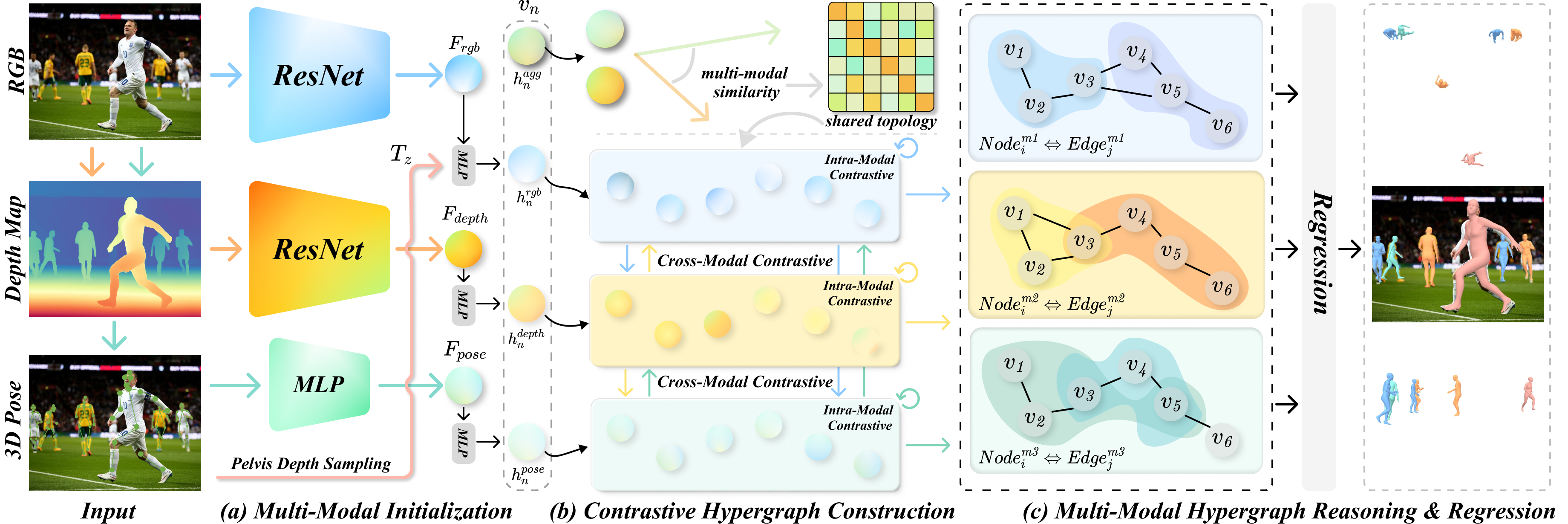}
    \vspace{-0.6em}

    \caption{\textbf{Overview of CoMHR.} We begin with \textit{(a) Multi-Modal Initialization}, which lifts 2D keypoints via pseudo-depth maps and explicitly anchors RGB features with Pelvis Depth to initialize multi-modal features. These cues are fused to construct \textit{(b) a Contrastive Hypergraph}, where a shared topology is regularized by contrastive learning. This structure enables \textit{(c) High-Order Reasoning} to propagate collective context, driving precise SMPL regression.}
    \label{fig:pipeline}
    \vspace{-1.0em}
\end{figure*}

\section{Method}\label{sec:Method}
In this work, we present a Multimodal Hypergraph Reasoning framework to mitigate occlusion and depth ambiguity in 3D reconstruction. Our approach begins by synergizing RGB, depth, and pose features to initialize robust node representations~(\secref{sec:multimodal_feature}). To capture inter-person dependencies, we dynamically construct a hypergraph structure explicitly regularized by contrastive learning~(\secref{sec:HypergraphConstruction}). Finally, we perform high-order reasoning to achieve robust and precise regression of multi-person human meshes~(\secref{sec:reasoning_regression}).

\subsection{Multi-Modal Node Initialization}\label{sec:multimodal_feature}

To model complex inter-individual relationships, we define a hypergraph $\mathcal{G} = (\mathcal{V}, \mathcal{E})$ where the node set $\mathcal{V} = \{v_n\}_{n=1}^N$ represents the $N$ detected individuals. 
Since single-modal cues are unreliable in crowded scenes due to occlusions and geometric ambiguities, we mitigate this by extracting three complementary modalities for each node $v_n$: RGB images for rich semantics, depth maps for geometric ordering, and occlusion-aware 3D poses for skeletal priors (as in \figref{fig:pipeline} (a)).

\subsubsection{Visual Semantic Encoding}
Directly processing the entire scene image compromises the resolution of individuals. To address this, we extract a high-resolution patch for each person $n$ centered at $(c_x, c_y)_n$ with scale $s_n$. These patches are resized to a fixed dimension $\mathbb{R}^{H \times W \times 3}$ and fed into a ResNet backbone to obtain the visual feature embedding $F_{\text{rgb}} \in \mathbb{R}^{N \times 2D}$.

\subsubsection{Geometric Structure Encoding}
Since RGB features lack depth ordering, we leverage Depth Anything V2~\cite{Yang2024DepthAU} to generate a pseudo-depth map. We mirror the RGB preprocessing by extracting single-channel patches ($\mathbb{R}^{H \times W \times 1}$) using the same bounding boxes, followed by a separate ResNet encoder to yield the geometric feature embedding $F_{\text{depth}} \in \mathbb{R}^{N \times 2D}$, providing crucial cues for resolving depth ambiguity.

\subsubsection{Occlusion-Aware Pose Encoding}
While RGB and depth capture dense details, they lack explicit skeletal priors crucial for inferring invisible parts under severe occlusion. Hence, we construct an incomplete 3D pose representation for each subject to complement the visual embeddings. We estimate 2D joints $J^{2D}_n = \{(x_k, y_k, c_k)\}_{k=1}^K$ via OpenPose~\cite{Hidalgo2019OpenPoseW} and apply a visibility mask $m_k = \mathbb{1}(c_k > \tau)$ (with $\tau=0.5$) to filter low-confidence joints. For valid joints, we sample depth $z_k$ from the aligned depth map to form 3D vectors $p_k = (x_k, y_k, z_k)$. These joint sequences are then processed by a lightweight 1D convolutional encoder, where the visibility masks $m_k$ guide weighted average pooling over the joint dimension, yielding the final pose embedding $F_{\text{pose}} \in \mathbb{R}^{N \times 2D}$ that effectively captures local 3D structural cues.

\subsubsection{Pelvis-Guided Global Spatial Embedding}
Beyond local structural constraints, we further leverage the pelvis depth $T_z$ as a global spatial anchor to explicitly model the individual's relative spatial arrangement in the scene. $T_z$ is computed as the average depth of the visible left and right hip joints: $T_z = (z_{\text{l\_hip}} + z_{\text{r\_hip}})/2$. Unlike local joint features, $T_z$ serves as a consistent anchor for determining relative depth ordering among multiple people. Considering that the depth and pose modalities inherently possess 3D geometric information, we exclusively concatenate this global anchor with the RGB features to compensate for their lack of spatial awareness.

\subsubsection{Nodes Initialization}
Finally, we map the extracted features into a unified latent space $\mathbb{R}^D$ via independent MLPs. This yields three sets of single-modal embeddings: $\mathbf{h}_n^{\text{rgb}}$ (augmented with $T_z$), $\mathbf{h}_n^{\text{depth}}$, and $\mathbf{h}_n^{\text{pose}}$. To capture holistic information, we further concatenate them into $\mathbf{h}_n^{\text{agg}}$:
\begin{equation}
    \mathbf{h}_n^{\text{agg}} = \left[ \mathbf{h}_n^{\text{rgb}} \mathbin{\|} \mathbf{h}_n^{\text{depth}} \mathbin{\|} \mathbf{h}_n^{\text{pose}} \right].
    \label{eq:node_init}
\end{equation}
These embeddings serve as the input for the subsequent construction of hypergraphs and the shared global topology that supports multi-modal contrastive learning and reasoning.

\subsection{Contrastive Hypergraph Construction}\label{sec:HypergraphConstruction}

To further exploit collective behaviors to compensate for incomplete individual observations, we propose to construct a \textit{Contrastive Hypergraph} as shown in \figref{fig:pipeline} (b). Instead of relying on pre-defined topologies, this module models high-order correlations by leveraging multi-modal complementarity via a contrastive learning objective. This explicitly enforces cross-modal consistency, ensuring robust graph construction even when individual modalities are corrupted.

\subsubsection{Dynamic Hyperedge Generation via Shared Topology}
To ensure structural consistency against single-modal noise, we infer a shared hypergraph topology from aggregated features $\mathbf{h}^{\text{agg}}$ (Eq.~\ref{eq:node_init}), rather than partial cues. First, we compute a global affinity matrix $\mathbf{A} \in \mathbb{R}^{N \times N}$ via normalized correlation between individual features:
\begin{equation}
    \mathbf{A}_{i,j} = \frac{(\mathbf{h}_i^{\text{agg}})^\top \mathbf{h}_j^{\text{agg}}}{\|\mathbf{h}_i^{\text{agg}}\|_2 \|\mathbf{h}_j^{\text{agg}}\|_2}.
\end{equation}
The element $\mathbf{A}_{i,j}$ quantifies the comprehensive similarity (semantic, geometric, and structural) between the $i$-th and $j$-th individuals.
To capture group-wise relationships, we define a hyperedge $e_i$ centered at each node $v_i$. Specifically, we identify the $K$ nearest neighbors of $v_i$ based on the global affinity scores in $\mathbf{A}$ to form a high-density sub-group.
The shared incidence matrix $\mathbf{H} \in \mathbb{R}^{N \times N}$ is then constructed as:
\begin{equation}
    \mathbf{H}_{j, i} = 
    \begin{cases} 
    1, & \text{if } v_j \in \mathcal{N}_K(v_i), \\
    0, & \text{otherwise},
    \end{cases}
\end{equation}
where $\mathcal{N}_K(v_i)$ denotes the neighbor set retrieved from $\mathbf{h}^{\text{agg}}$. 
This generated structure $\mathbf{H}$ is applied to all modalities, ensuring message passing in the RGB, depth, and pose branches with a consistent and reliable high-order topology.

\subsubsection{Hypergraph Contrastive Learning}

The quality of the generated shared hypergraph hinges on the discriminativeness of the node features. Since the shared topology is derived from the aggregated features $\mathbf{h}^{\text{agg}}$, it is imperative that each constituent modality provides reliable and distinct cues. To achieve this, we introduce a dual-branch contrastive learning strategy to regularize the feature space before aggregation.

\noindent\textbf{Intra-Modal Contrastive Learning.}
To refine single-modal representations, we treat nodes with similar ground-truth 3D poses (low MPJPE) as positive pairs $\mathcal{P}(i)$. We minimize the intra-modal loss to enforce tighter semantic clustering among these positive pairs and enhance the structural coherence of single-modal features:
\begin{equation}
    \mathcal{L}_{\text{intra}}^{(m)} = - \sum_{i \in \mathcal{V}} \frac{1}{|\mathcal{P}(i)|} \sum_{p \in \mathcal{P}(i)} \log \frac{\exp(S_{i,p})}{\sum_{p' \in \mathcal{P}(i)} \exp(S_{i,p'})},
\end{equation}
with scaled similarity $S_{i,j} = \text{sim}(\mathbf{h}_i^{(m)}, \mathbf{h}_j^{(m)}) / \tau$.

\noindent\textbf{Cross-Modal Contrastive Learning.}
Since $\mathbf{h}^{\text{agg}}$ is a concatenation, inter-modal redundancy limits representation capacity. To enhance complementarity, we introduce an orthogonality constraint that explicitly regulates the cosine similarity between distinct modalities of the same node. To this end, we apply a negative average cosine similarity loss with ReLU activation:
\begin{equation}
    \mathcal{L}_{\text{cross}} = \frac{1}{|\mathcal{V}|} \!\sum_{i \in \mathcal{V}} \!\max \Big(0, -1/3 \!\!\!\sum_{m_1 < m_2} \!\!\!\cos(\mathbf{h}_i^{(m_1)}, \!\mathbf{h}_i^{(m_2)}) \Big)
\end{equation}
This formulation penalizes negative correlations to ensure that heterogeneous modalities capture distinct and complementary traits, thereby maximizing the information entropy in $\mathbf{h}^{\text{agg}}$.
The total contrastive loss is a weighted sum: $\mathcal{L}_{\text{contrastive}} = \sum_{m} \mathcal{L}_{\text{intra}}^{(m)} + \alpha \mathcal{L}_{\text{cross}}$, where $\alpha$ is a balancing weight, and the objective is jointly optimized with regression losses.

\subsection{Multi-Modal Hypergraph Reasoning and Regression}\label{sec:reasoning_regression}

Leveraging the shared topology $\mathbf{H}$, we perform high-order reasoning via message passing to dynamically exploit the complementarity of semantic, geometric, and pose priors, as shown in \figref{fig:pipeline} (c). This unified propagation mitigates single-modal fragility (\eg, occlusion) by utilizing collective neighbor cues to rectify local noise, enforcing a consistent global context across all branches.
For each modality $m$, the feature update is performed via a two-stage mechanism:

\subsubsection{Node-to-Hyperedge Aggregation}
We first aggregate information from individuals within hyperedge to form group-level representations. For each hyperedge $e_j$, the group feature $\mathbf{f}_j^{(m)}$ is computed by aggregating its constituent node features:
\begin{equation}
    \mathbf{f}_j^{(m)} = \sigma \left( \sum_{v_i \in e_j} \mathbf{h}_i^{(m)} \mathbf{W}_{\text{agg}}^{(m)} \right),
\end{equation}
where $\mathbf{W}_{\text{agg}}^{(m)}$ is the modality-specific aggregation weight and $\sigma$ is a non-linear activation. This step extracts collective patterns (\eg, shared motion or spatial arrangement) from the group.

\begin{figure*}[t]
    \centering
    \includegraphics[width=0.95\linewidth]{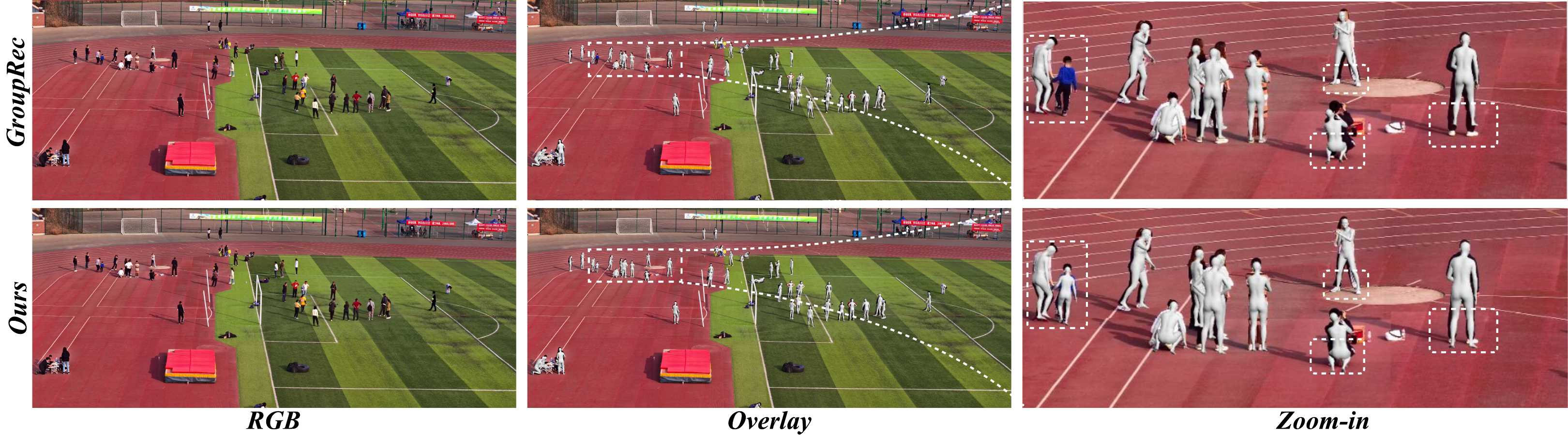}
    \vspace{-0.6em}
    \caption{\textbf{Qualitative comparison on GigaCrowd~\cite{Li2021DeepSG}}.
GroupRec~\cite{Huang2023ReconstructingGO} often shows floating feet, drifting bodies, and incorrect depth ordering, while our method maintains accurate depth, stable spatial layout, and consistent reconstruction in dense crowds.
    }
    \label{fig:giga_quad}
    \vspace{-1.0em}
\end{figure*}

\subsubsection{Hyperedge-to-Node Update}
Next, group features are propagated back to refine the nodes. Each node $v_i$ is updated by aggregating context from its connected hyperedges:
\begin{equation}
    \tilde{\mathbf{h}}_i^{(m)} = \sigma \left( \sum_{e_j \ni v_i} \mathbf{f}_j^{(m)} \mathbf{W}_{\text{update}}^{(m)} \right).
\end{equation}
This mechanism utilizes group context to refine occluded features while retaining unique modal characteristics.

\subsubsection{Human Parameters Regression}
To exploit multi-modal complementarity, we concatenate the refined features with the bounding box embedding $b_n$ (encoding scale following~\cite{Li2022CLIFFCL}):
\begin{equation}
    \mathbf{v}_n^{\text{final}} = \left[ \tilde{\mathbf{h}}_n^{\text{rgb}} \mathbin{\|} \tilde{\mathbf{h}}_n^{\text{depth}} \mathbin{\|} \tilde{\mathbf{h}}_n^{\text{pose}} \mathbin{\|} \phi_{\text{box}}(b_n) \right].
\end{equation}
This vector is then fed into the SMPL regressor to predict pose $\theta$, shape $\beta$, and camera parameters, integrating high-order structural reasoning with rich multi-modal evidence.

To train the network in an end-to-end manner, we employ a multi objective function that enforces geometric accuracy and feature discriminativeness simultaneously. We supervise the regression using losses on 2D reprojection ($\mathcal{L}_{\text{reproj}}$), SMPL parameters ($\mathcal{L}_{\text{smpl}}$), and 3D joint coordinates ($\mathcal{L}_{\text{joint}}$). Incorporating the feature regularization term ($\mathcal{L}_{\text{contrastive}}$) defined in Sec.~\ref{sec:HypergraphConstruction}, the total loss is formulated as:
\begin{equation}
    \mathcal{L}_{\text{total}} = \lambda_1 \mathcal{L}_{\text{reproj}} + \lambda_2 \mathcal{L}_{\text{smpl}} + \lambda_3 \mathcal{L}_{\text{joint}} + \lambda_4 \mathcal{L}_{\text{contrastive}},
\end{equation}
where $\lambda_{\{1,2,3,4\}}$ are weights to balance each component.

\section{Experiments}\label{sec:Experiments}
\subsection{Experimental Setup}
\subsubsection{Datasets}\label{sec:datasets}
We train on COCO~\cite{Lin2014MicrosoftCC} and MPII~\cite{Andriluka20142DHP} using 2D and CLIFF-generated pseudo-3D~\cite{Li2022CLIFFCL} annotations to supervise 3D losses and contrastive learning. For quantitative evaluation, we employ Panoptic Studio~\cite{Joo2015PanopticSA} and GigaCrowd~\cite{Li2021DeepSG} for accuracy and large-scale crowd assessment. Additionally, we provide qualitative visualizations on CrowdPose~\cite{Li2018CrowdPoseEC} under varied crowded scenarios.

\subsubsection{Metrics}\label{sec:metrics} 
Following~\cite{Huang2023ReconstructingGO}, we adopt pose-level and group-level metrics. At the individual level, we report Mean Per Joint Position Error (MPJPE). At the group level, we evaluate on GigaCrowd using \textbf{OKS}, \textbf{PA-PPDS}, \textbf{PCOD}, and \textbf{RP}. Higher OKS, PA-PPDS, and PCOD indicate better consistency, while lower RP denotes fewer reconstruction conflicts.
\begin{figure*}[t]
    \centering
    \includegraphics[width=0.95\textwidth]{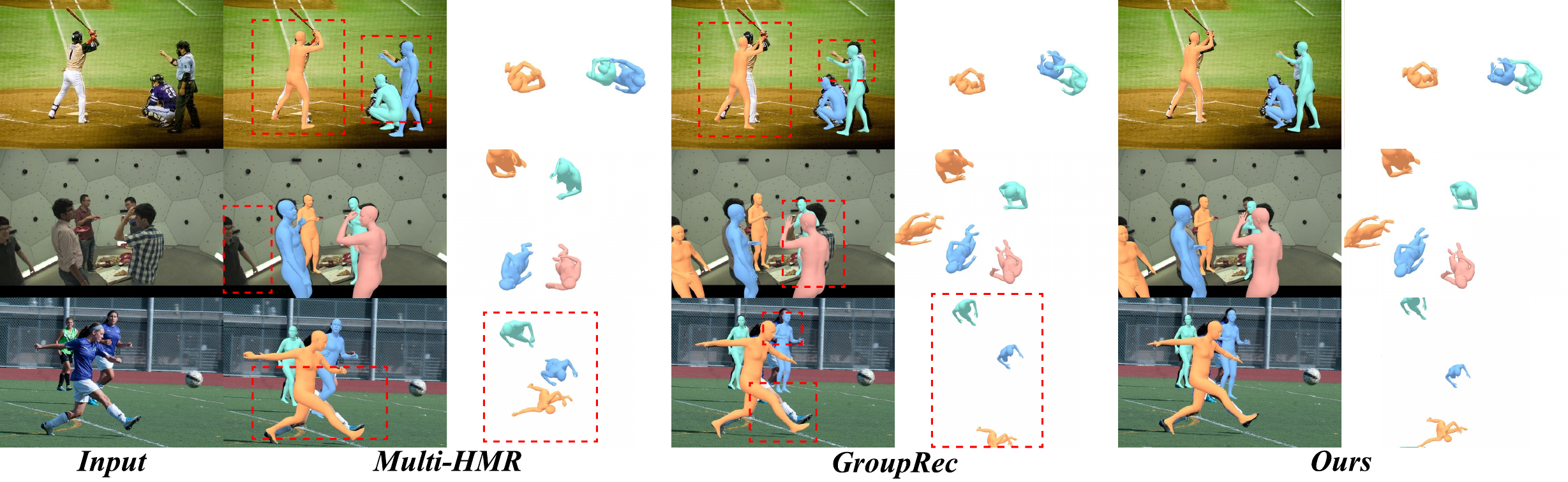}
    \vspace{-0.6em}
    \caption{
    \textbf{Qualitative comparison of multi-person reconstruction} among Multi-HMR~\cite{Baradel2024MultiHMRMW}, GroupRec~\cite{Huang2023ReconstructingGO}, and our method.
    Red boxes show baseline failures such as pose misalignment, missing people, and depth errors, while our method produces accurate and consistent reconstructions.
    }
    \label{fig:comparison}
    \vspace{-1.0em}
\end{figure*}

\begin{table}[t]
    \centering
    \caption{Comparison with multi-person mesh recovery methods on Panoptic dataset. MPJPE (mm); '--' indicates unavailable results. Best in \textbf{bold}, second best \underline{underlined}.}
    \setlength{\tabcolsep}{4.5pt}
    \begin{tabular}{p{1.8cm}ccccc} 
        \toprule
        \textbf{Method} & \textbf{Haggling} & \textbf{Mafia} & \textbf{Ultim.} & \textbf{Pizza} & \textbf{Mean} \\
        \midrule
        Zanfir~\textit{et~al.}~\cite{Zanfir2018Monocular3P} & 140.0 & 165.9 & 150.7 & 156.0 & 153.4 \\
        MubyNet~\cite{Zanfir2018DeepNF} & 141.4 & 152.3 & 145.0 & 162.5 & 150.3 \\
        CRMH~\cite{Jiang2020CoherentRO} & 129.6 & 133.5 & 153.0 & 156.7 & 143.2 \\
        BMP~\cite{Zhang2021BodyMA} & 120.4 & 132.7 & 140.9 & 147.5 & 135.4 \\
        Pose2UV~\cite{Huang2022Pose2UVSM} & 104.2 & 136.0 & 123.2 & 151.0 & 128.6 \\
        ROMP~\cite{Sun2020MonocularOR} & 110.8 & 122.8 & 141.6 & 137.6 & 128.2 \\
        3DCrowdNet~\cite{Choi2021LearningTE} & 109.6 & 135.9 & 129.8 & 135.6 & 127.3 \\       Luvizon~\textit{et~al.}~\cite{Luvizon2023SceneAware3M} & 93.6 & -- & 133.8 & 145.9 & -- \\
        SMAP~\cite{Zhen2020SMAPSM} & 128.5 & - &141.2 &236.4 &168.7\\
        Crowd3D~\cite{Wen2023Crowd3DTH} & 100.1 & - & 125.7 & 134.1 & 120.0 \\
        BEV~\cite{Sun2021PuttingPI} & 90.7 & \underline{103.7} & \underline{113.1} & 125.2 & 109.5\\        GroupRec~\cite{Huang2023ReconstructingGO} & \underline{86.8} & 107.8 & \textbf{110.7} & \underline{121.1} & \underline{106.6} \\
        \midrule
        \textbf{Ours} & \textbf{86.0} & \textbf{99.5} & 113.6 & \textbf{117.6} & \textbf{104.2} \\
        \bottomrule
    \end{tabular}%
    \label{tab:panoptic_comparison}
    \vspace{-2em}
\end{table}

\subsection{Comparison with State-of-the-Art Methods}\label{sec:Comparison}
We evaluate our method on the GigaCrowd dataset, featuring dense crowds with large scale variation, heavy occlusions, and complex layouts. Qualitative results in \figref{fig:giga_quad} show that GroupRec~\cite{Huang2023ReconstructingGO} suffers from floating feet, drifting bodies, and inconsistent depth. By leveraging multi-modal depth cues and hypergraph reasoning, our method accurately recovers metric-scale-agnostic depth ordering and spatial layout, producing stable reconstructions even in crowded scenes with hundreds of people. As shown in \tabref{tab:gigacrowd_comparison}, our approach outperforms state-of-the-art methods across all metrics. Notably, recent one-stage models like SAT-HMR~\cite{Su_2025_CVPR} fail on gigapixel images due to catastrophic pixel loss during downscaling, causing performance collapse (0.79 OKS, 4.04 PA-PPDS). Unlike such methods, CoMHR processes original resolutions without information loss, ensuring robust estimation and establishing superiority over GroupRec. Compared to GroupRec, pose consistency (OKS) improves by 7.3\%, cross-individual spatial consistency (PA-PPDS) by 8.5\%, reconstruction conflicts are eliminated (RP = 0.00), and PCOD increases by 22.9\%, enhancing mesh alignment.

We further present representative multi-person scenes from CrowdPose~\cite{Li2018CrowdPoseEC} and Panoptic~\cite{Joo2015PanopticSA}, comparing our approach with Multi-HMR~\cite{Jiang2020CoherentRO} and GroupRec~\cite{Huang2023ReconstructingGO} in \figref{fig:comparison}. Red boxes highlight typical failure regions of existing methods, including local pose misalignments, missed detections, and misaligned 2D projections, especially in areas with heavy occlusion or overlapping interactions. In contrast, our approach leverages multi-modal feature fusion and high-order hypergraph relational modeling to propagate global context, maintaining stable reconstructions even in dense, interactive scenes. Consequently, our model accurately recovers depth ordering, preserves pose coherence, and demonstrates strong spatial consistency while robustly handling occlusions.

\begin{table}[t]
    \centering
    \caption{Comparison with state-of-the-art methods on GigaCrowd.}
    \begin{tabular}{lcccc}
        \toprule
        \multirow{2}{*}{\textbf{Method}} & \textbf{OKS} & \textbf{PA-PPDS} & \textbf{PCOD} & \textbf{RP} \\
         & \scriptsize{(\%) $\uparrow$} & \scriptsize{(\%) $\uparrow$} & \scriptsize{(\%) $\uparrow$} & \scriptsize{$\downarrow$} \\
        \midrule
        CRMH~\cite{Jiang2020CoherentRO} & 56.31 & 52.16 & 60.48 & 0.17\\
        BEV~\cite{Sun2021PuttingPI} & 62.47 & 55.41 & 62.38 & 0.22 \\
        CrowdRec~\cite{Huang2023CrowdRec3C} & 65.65 & 73.83 & \underline{81.69} &  \textbf{0.00} \\
        GroupRec~\cite{Huang2023ReconstructingGO} & \underline{70.80} & \underline{67.22} & 71.42 & 0.17 \\
        \midrule
        \textbf{Ours} & \textbf{78.11} & \textbf{75.74} & \textbf{94.35} & \textbf{0.00} \\
        \bottomrule
    \end{tabular}
    \label{tab:gigacrowd_comparison}
    \vspace{-1em}
\end{table}
Quantitative comparisons on Panoptic~\cite{Joo2015PanopticSA} are summarized in \tabref{tab:panoptic_comparison}. Our method achieves an average MPJPE of 104.2 mm, outperforming CRMH, ROMP, GroupRec, and BEV. This demonstrates that contrastive learning-enhanced multi-modal feature fusion strengthens intra-modal discriminability and promotes cross-modal complementarity, enabling reliable reconstruction under varying motion complexity and occlusions. Across individual sequences, our method demonstrates superior performance. On Haggling and Mafia, it achieves 86.0 mm and 99.5 mm, respectively, compared to GroupRec’s 86.8 mm and 107.8 mm, corresponding to improvements of 0.8 mm and 8.3 mm. In the most challenging Pizza sequence, our method reduces errors by 7.6 mm and 20.0 mm compared to BEV and ROMP, respectively. Overall, the average standard deviation across sequences is 11.2 mm, indicating model stability and consistency under diverse motions and occlusion.

\subsection{Ablation Study}\label{sec:Ablation}
We conduct ablation studies on the Panoptic~\cite{Joo2015PanopticSA} dataset to evaluate the contributions of multimodal features, translation refinement, and contrastive learning to multi-person 3D reconstruction accuracy (\tabref{tab:ablation_panoptic}). Single-modal RGB achieves an MPJPE of 110.41 mm, outperforming depth-only and pose-only variants, all of which rely solely on ResNet features and direct pose regression without hypergraph reasoning. Introducing a second modality (RGB+Depth or RGB+Pose) activates hypergraph relational modeling and substantially reduces the error. Using all three modalities yields further improvements, and incorporating translation refinement provides an additional accuracy gain. Finally, adding contrastive learning achieves the best MPJPE of 104.18 mm by strengthening cross-modal alignment and enhancing robustness under heavy occlusion. 
Overall, the results demonstrate that multimodal fusion, hypergraph reasoning, translation refinement, and contrastive learning collectively and significantly improve both the accuracy and stability of multi-person 3D reconstruction.

\begin{table}[t]
    \centering
    \caption{
    \textbf{Ablation study} on the \textbf{Panoptic} dataset.
    ``Tz'' denotes translation refinement from depth normalization.
    }
    \renewcommand{\arraystretch}{1.2}
    \setlength{\tabcolsep}{4pt}
    \begin{tabular}{p{3.1cm}cc} 
        \toprule
        \textbf{Configuration} & \textbf{Contrastive Learning} & \textbf{MPJPE} \scriptsize{(mm) $\downarrow$} \\
        \midrule
        RGB only & $\times$ & 110.41 \\
        Depth only & $\times$ & 137.40 \\
        Pose only & $\times$ & 184.36 \\
        RGB + Depth & $\times$ & 106.90 \\
        RGB + Pose & $\times$ & 106.89 \\
        RGB + Depth + Pose & $\times$ & 106.34 \\
        RGB + Depth + Pose (Tz) & $\times$ & 105.26 \\
        RGB + Depth + Pose (Tz) & \checkmark & \textbf{104.18} \\
        \bottomrule
    \end{tabular}
    \label{tab:ablation_panoptic}
    \vspace{-2em}
\end{table}

\section{Conclusion}\label{sec:Conclusion}
We propose a multi-modal hypergraph reasoning framework for multi-person 3D reconstruction in complex scenes. By fusing RGB, depth, and 3D pose features and explicitly modeling high-order inter-person relations with intra-modal and cross-modal contrastive learning, the method achieves robust reconstruction under occlusion and dense interactions. Experiments and ablation studies demonstrate that multi-modal feature fusion and hypergraph reasoning significantly improve reconstruction accuracy and spatial consistency.

\bibliographystyle{IEEEtran}
\bibliography{main}

\clearpage
\section*{\LARGE Supplementary Material}
\addcontentsline{toc}{section}{Supplementary Material}
In this document, we provide the following supplementary contents to further clarify the proposed method and facilitate reproducibility:
In this document, we provide the following supplementary contents to further clarify the proposed method and facilitate reproducibility:
\begin{itemize}
    \item \textbf{1. Extended Ablation Studies:} In-depth analysis of crowd-level metrics and step-by-step visual improvements for robust reconstruction.
    \item \textbf{2. Additional Qualitative Results:} Extended visualizations on CrowdPose and GigaCrowd.
    \item \textbf{3. Implementation Details:} Network architecture, hyperparameters, and hardware setups.
    \item \textbf{4. Robustness to Noise and Upstream Failures:} Stress tests against environmental and model corruptions.
    \item \textbf{5. Efficiency and Scalability:} Parameter efficiency, inference speed, and $\mathcal{O}(M)$ complexity.
    \item \textbf{6. Limitations:} Discussions on extreme failure cases, foundation model dependencies, and future directions.
    \item \textbf{7. Multimodal Contrastive Learning Pseudocode:} Algorithmic formulation of our loss mechanisms.
\end{itemize}

\subsection*{1. Extended Ablation Studies}
In this section, we present ablation studies to further validate the individual contributions of our proposed modules. Specifically, we provide quantitative evaluations on crowd-level spatial metrics and qualitative visualizations to demonstrate step-by-step improvements of our CoMHR framework.

\begin{figure*}[b]
    \centering
    \includegraphics[width=0.9\linewidth]{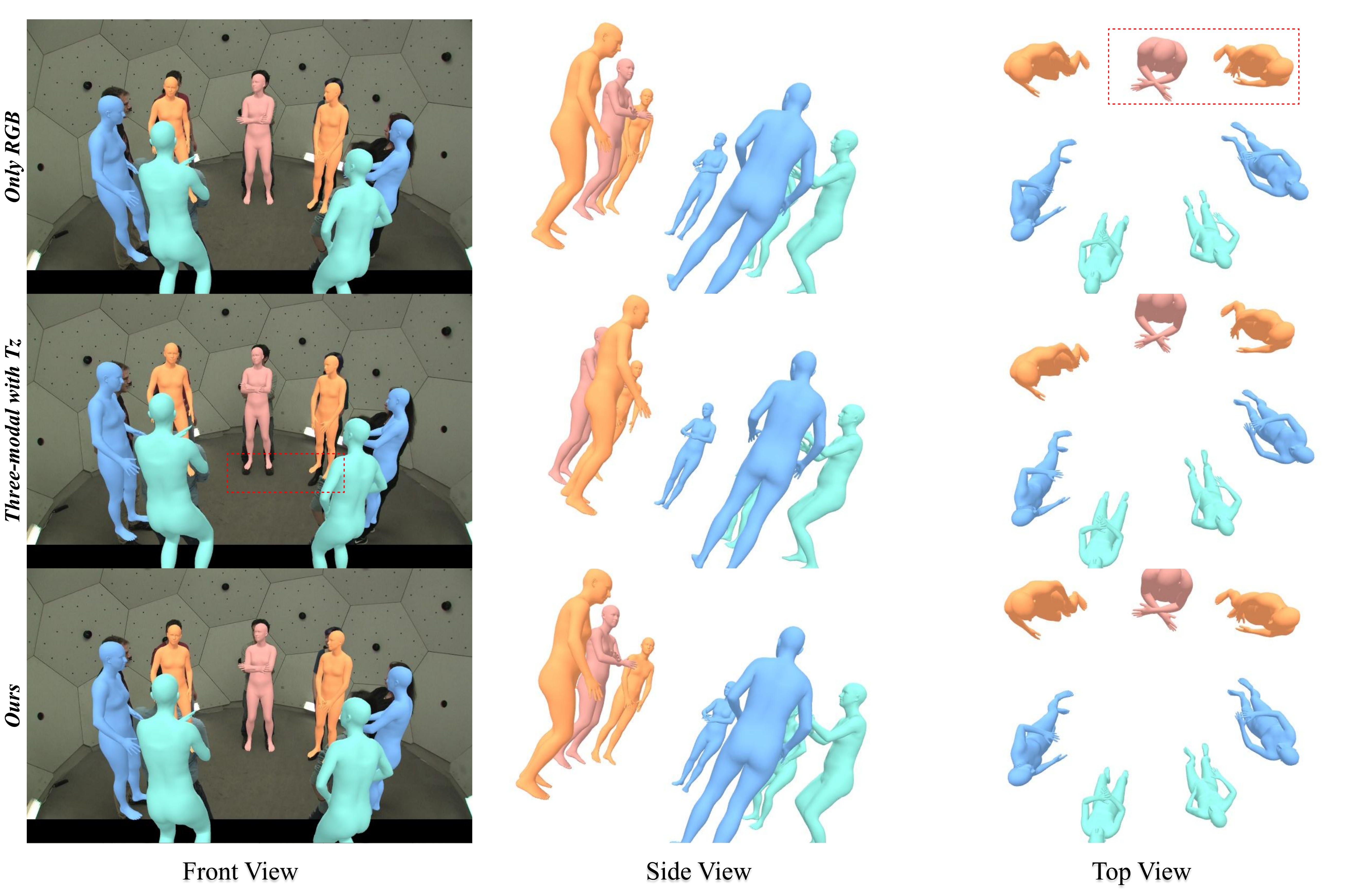}
    \vspace{-1em}
    \caption{\textbf{Ablation results on the Panoptic dataset.}
    Top to bottom: (1) RGB-only baseline with depth errors, (2) RGB+Depth+Pose+$T_z$ with improved 3D structure, (3) full model with contrastive learning achieving accurate depth and coherent meshes.}
    \label{fig:sup_ablation_analysis}
    \vspace{-1.5em}
\end{figure*}
\addcontentsline{toc}{subsection}{1. Extended Ablation Studies}
\noindent\textbf{Quantitative Analysis on Crowd-Level Metrics.}
Beyond individual MPJPE, we further evaluate our method on crowd-level metrics using the GigaCrowd dataset, as shown in Tab.~\ref{tab:metrics}. The results demonstrate that contrastive learning yields substantial gains in OKS, PA-PPDS, and PCOD, directly proving the effectiveness of our topological modeling over mere reliance on foundation model priors. Furthermore, we observe that employing an early-fusion strategy with shared encoder weights causes the MPJPE to increase by 5.1 mm, as heterogeneous feature spaces in crowds easily lead to cross-modal contamination. Our multi-modal hypergraph design mitigates this by explicitly enforcing cross-modal orthogonality. Finally, CoMHR exhibits strong insensitivity to hyperparameter variations, with performance fluctuating less than 3.6 mm when $\tau \in [0.05, 0.08]$, $\epsilon \in [0.1, 0.2]$, and the loss weight $\alpha \in [0.03, 0.05]$.

\vspace{-1em}
\begin{table}[h]
\centering
\caption{Ablation on Crowd-Level Metrics on GigaCrowd.}
\label{tab:metrics}
    \resizebox{0.48\textwidth}{!}{
    \begin{tabular}{lcccc}
        \toprule
        Configuration & OKS $\uparrow$ & PA-PPDS $\uparrow$ & PCOD $\uparrow$ & RP $\downarrow$ \\
        \midrule
        RGB + Pose Baseline & 72.93 & 75.27 & 94.03 & \textbf{0.0} \\
        CoMHR w/o Contrastive & 69.38 & \textbf{76.11} & 92.98 & \textbf{0.0} \\
        \textbf{Full CoMHR} & \textbf{78.11} & 75.74 & \textbf{94.35} & \textbf{0.0} \\
        \bottomrule
    \end{tabular}
    }
    \vspace{-1em}
\end{table}
Consistent with the quantitative findings, we present qualitative ablation results for three variants of our model: the RGB-only baseline, a multimodal variant incorporating RGB, depth, pose, and pelvis-centered depth cues ($T_z$), and the full model further augmented with contrastive learning. As shown in \figref{fig:sup_ablation_analysis}, the RGB baseline is able to capture 2D appearance and body silhouettes well, generating complete and natural human meshes without physically implausible artifacts such as disconnected or floating limbs. However, due to the lack of explicit geometric constraints, the predicted depth of individuals does not always align with the actual scene, resulting in inaccurate global depth ordering, especially in densely crowded or heavily occluded scenarios.

The multimodal variant, which integrates depth, pose, and pelvis-centered depth cues, provides stronger and more explicit geometric guidance, enabling the network to reason more accurately about the 3D human structure. By leveraging complementary cues from multiple modalities, this variant effectively alleviates depth ambiguities commonly encountered in monocular RGB-based reconstruction, producing human meshes with improved global depth ordering, more coherent inter-person spatial relationships, and consistent body proportions. Although 2D projection alignment may be slightly compromised due to the network balancing multiple modalities, the overall 3D reconstruction becomes substantially more reliable, with meshes remaining complete and physically plausible. 

Incorporating the contrastive learning module further improves the model by explicitly aligning feature representations across modalities in the latent space. This structured alignment encourages RGB, depth, and pose embeddings to share a coherent and discriminative latent space, facilitating more effective cross-modal feature fusion. Consequently, the full model achieves both accurate depth estimation and precise 2D projection consistency, resulting in higher-quality and more robust mesh reconstructions. The advantages of the complete model are especially evident in challenging scenarios with dense crowds, severe occlusions, and diverse human poses. Overall, this comparison clearly demonstrates the importance of multimodal features and contrastive alignment in improving 3D reconstruction quality, stability, and spatial coherence, particularly in complex, real-world crowd scenarios.

\begin{figure}[t]  
    \centering
    \includegraphics[width=0.9\linewidth]{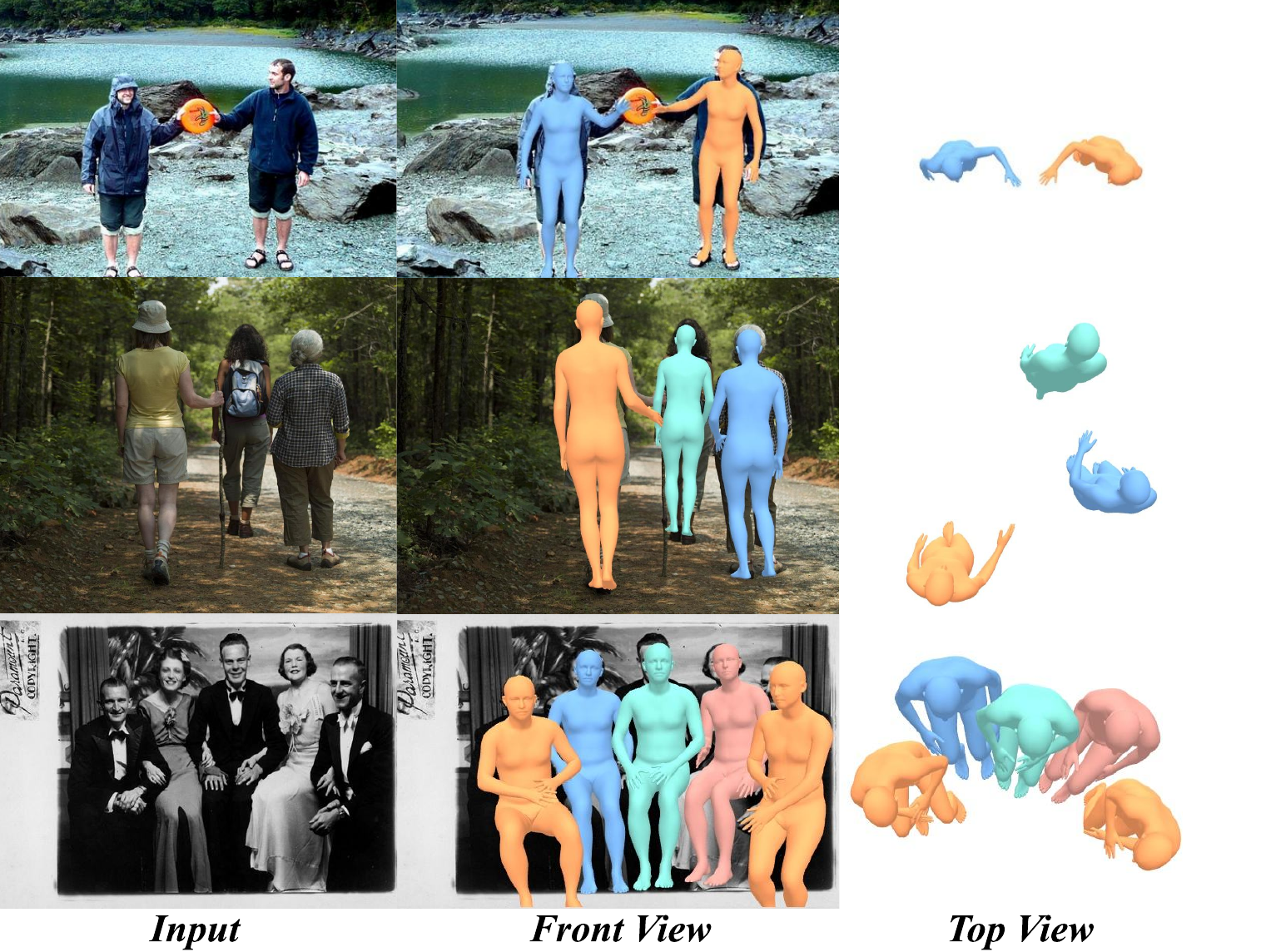}
    \vspace{-0.8em}
    \caption{Additional qualitative results showing high-order interactions and relative depth ordering.}
    \label{fig:teaser_appendix}
    \vspace{-1em}
\end{figure}

\subsection*{2. Additional Qualitative Results} 
\addcontentsline{toc}{subsection}{2. Additional Qualitative Results} 

To further demonstrate our method's robustness and versatility across diverse datasets and crowd densities, \figref{fig:teaser_appendix} shows three representative examples from the CrowdPose dataset, which is known for its crowded scenes, frequent inter-person occlusions, and challenging pose variations. Despite these difficulties, our model reconstructs coherent and physically plausible human meshes across a wide range of poses and interactions. The predicted meshes accurately capture fine-grained local joint articulations as well as global body configurations, exhibiting strong alignment with the visual evidence in the input images. Moreover, the reconstructions maintain consistent human proportions across individuals, demonstrating the model’s ability to handle diverse body shapes and poses. 

\begin{figure*}[t]
    \centering
    \includegraphics[width=\linewidth]{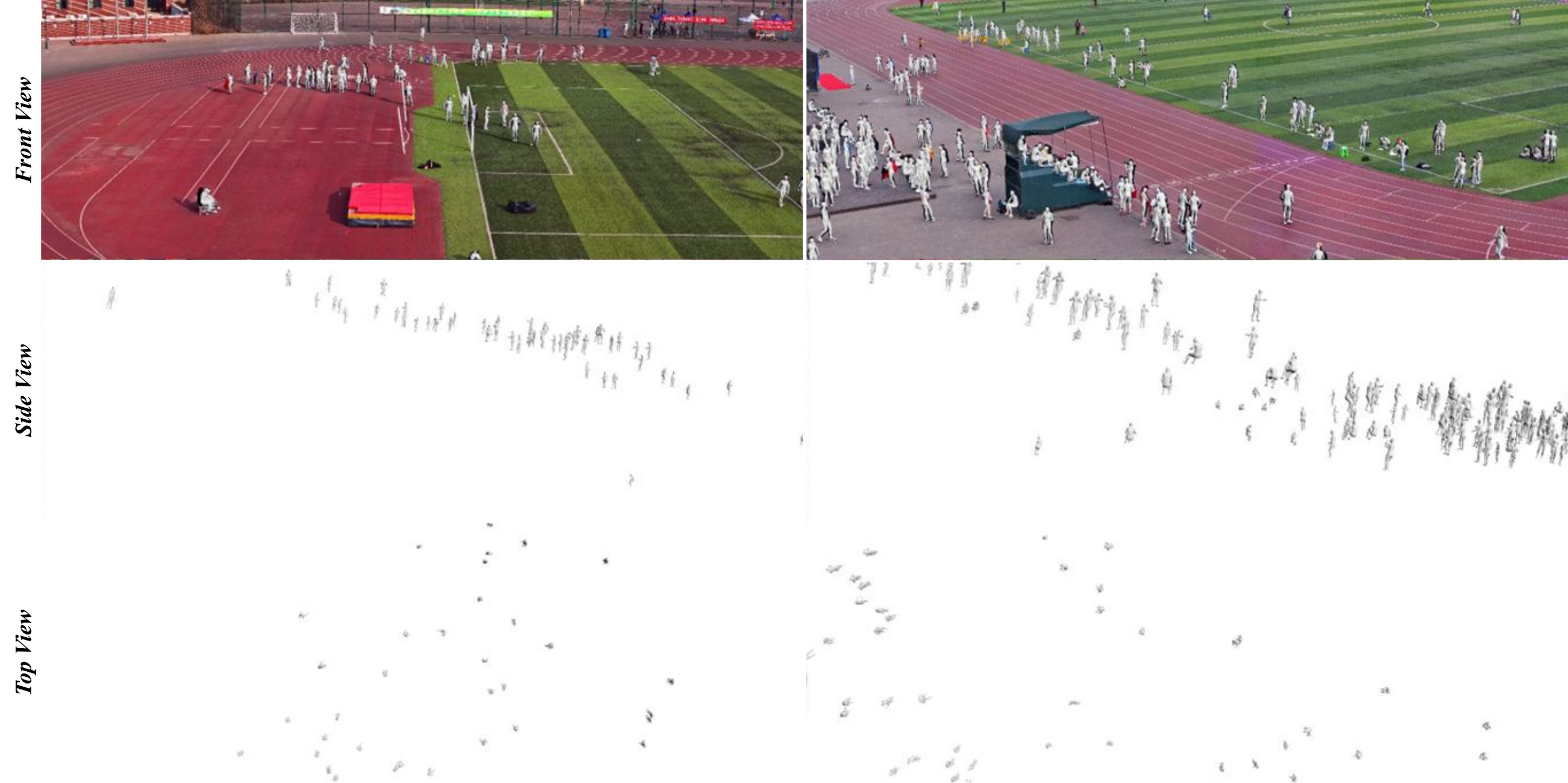}
    \vspace{-2em}
    \caption{\textbf{Multi-view results on GigaCrowd.}
    Two scenes are displayed, each showing front, side, and top views to illustrate the model’s ability to handle dense crowds and maintain depth consistency.}
    \label{fig：GigaSupple}
    \vspace{-1em}
\end{figure*}

\figref{fig：GigaSupple} presents qualitative results on the GigaCrowd dataset, which contains extremely dense, large-scale crowd scenes with severe occlusions and complex spatial arrangements. For each scene, we visualize the reconstructed meshes from multiple viewpoints, including front, side, and top-down perspectives, to better illustrate the 3D structure. Even in cases where numerous individuals heavily overlap in the image space, our model preserves correct global and relative depth relationships, avoids interpenetrations, and maintains stable and coherent mesh geometries. These reconstructions also demonstrate reliable spatial consistency between adjacent people, capturing realistic inter-person distances and interactions.

Together, these additional qualitative examples highlight the strong generalization capability of our approach across different datasets, crowd densities, and occlusion levels. The consistent reconstruction quality under challenging conditions confirms that the multimodal design effectively leverages RGB, depth, and pose cues to enforce geometric consistency and spatial coherence. This robustness makes the proposed method well-suited for real-world applications involving dense crowds, complex human interactions, and scenarios where traditional monocular reconstruction methods fail.

\subsection*{3. Implementation Details}
\addcontentsline{toc}{subsection}{3. Implementation Details}

To facilitate future research and ensure the full reproducibility of our work, we provide comprehensive implementation details of the proposed CoMHR framework in this section. Specifically, we elaborate on the hardware environment and training hyperparameters, the precise dimensions of our network architecture, and the exact configurations of our contrastive learning modules for stable convergence.

\vspace{0.5em}
\noindent\textbf{(a) Experimental Environment and Training Settings}

All experiments were conducted on a high-performance workstation equipped with a single NVIDIA RTX~4090 GPU (24\,GB VRAM), 120\,GB of system RAM, and a 16-core Intel Xeon Platinum~8352V CPU, running Python~3.10 and PyTorch~2.1.2 with CUDA~11.8 on Ubuntu~22.04. This hardware configuration allowed efficient processing of high-resolution images and large batch sizes while minimizing memory bottlenecks, which is critical for training a multimodal network that integrates RGB, depth, and pose features.

The model was trained for 60 epochs with a batch size of 32. The initial learning rate was set to $1\times10^{-4}$ and was decayed using a cosine annealing schedule with restarts, which helps the optimizer escape shallow local minima and promotes more stable convergence. Eight data-loading worker threads were employed to enable parallelized data I/O, reducing training latency. For training, the COCO and MPII\_CLIFF datasets were used to provide diverse multi-person images with rich 2D and 3D pose annotations, while the Panoptic dataset served as a benchmark for evaluation, allowing us to assess model performance on complex and controlled scenes. Standard data augmentation techniques—including scaling, rotation, horizontal flipping, and color jittering—were applied to improve generalization, and depth/pose maps were carefully aligned with RGB images to ensure multimodal consistency.

\vspace{0.5em}
\noindent\textbf{(b) Model Architecture} 

Our CoMHR network predicts SMPL parameters from RGB, depth, and 3D pose cues through a unified multimodal architecture designed to integrate complementary information from all input modalities.

\begin{itemize}
    \item \textbf{Feature Dimensions:} Each modality encoder (RGB, depth, pose) produces a 2048-dimensional feature vector. These vectors, along with the 3-dimensional pelvis depth anchor ($T_z$), are concatenated into a 6147-dimensional fused representation, which captures both appearance and geometric cues necessary for downstream relational reasoning across multiple people in crowded scenes.
    \item \textbf{Projection Head:} The fused vector is first normalized using LayerNorm and then projected into a 256-dimensional latent space. This compact latent representation is used both for contrastive learning, to align cross-modal features, and for SMPL regression, providing a rich yet manageable embedding for prediction tasks.
    \item \textbf{Prediction Heads:} The network employs three lightweight, modality-agnostic heads: a 6D rotation-based pose head predicting 24 joints, a 10-dimensional shape head, and a 3-dimensional camera translation head. The shared latent embedding ensures that the network can leverage cross-modal information for more accurate and consistent predictions, improving spatial coherence across multiple people.
\end{itemize}

\vspace{0.5em}
\noindent\textbf{(c) Contrastive Learning}

\begin{itemize}
    \item \textbf{Intra-modal:} Within each modality, positive sample pairs are defined using MPJPE distances. An NCE-style softmax is applied to maximize similarity between positive pairs. Crucially, by explicitly excluding negative samples from the denominator via a masking mechanism, the objective tightly focuses on aligning positive pairs, encouraging the network to learn discriminative embeddings that respect intra-modal geometric relationships.
    \item \textbf{Cross-modal:} To enforce alignment across modalities, RGB, depth, and pose embeddings are projected into a shared latent space and optimized using an orthogonality constraint. A negative average cosine similarity loss with ReLU activation is applied to prevent positive correlations, and the cross-modal loss is scaled by $\alpha = 0.03$ to balance its influence against intra-modal losses, ensuring stable training and feature disentanglement.
\end{itemize}

\subsection*{4. Robustness to Noise and Upstream Failures} 
\addcontentsline{toc}{subsection}{4. Robustness to Noise and Upstream Failures}

To comprehensively evaluate the resilience of CoMHR, we conducted stress tests under both environmental degradation and extreme upstream foundation model failures.

\vspace{0.5em}
\noindent\textbf{Environmental Degradation.} We tested the model's resistance to severe environmental noise on the Panoptic dataset (Baseline MPJPE = 104.18 mm). Under Foreground Truncation, where individuals are partially masked ($s_i \sim U(0,0.1)$), the error increased by only 2.41 mm. Under Sensor Contamination, simulated via Gaussian noise ($\sigma=0.1$), the error rose by merely 1.18 mm, demonstrating strong robustness to sensor and occlusion noise.
    
\vspace{0.5em}
\noindent\textbf{Upstream Foundation Model Failures.} To evaluate the resilience against upstream model corruptions (\eg, Depth Anything, OpenPose), we conducted further stress tests. First, to simulate a pelvis detection failure, we zeroed out the central 60\% $\times$ 60\% of the depth maps to simulate missing hips. Our cross-modal architecture allowed the RGB and pose branches to successfully compensate, altering the error by merely +0.05 mm. Second, to simulate topology chaos, we injected severe $\mathcal{N}(0,(3\sigma)^{2})$ bias into the $T_z$ anchor with a probability of $p=30\%$. The performance remained virtually unaffected, with an error increase of only +0.002 mm. Because $T_z$ anchors the RGB branch, our isolated feature extraction mechanism effectively blocks cross-modal error propagation.

\subsection*{5. Efficiency and Scalability}
\addcontentsline{toc}{subsection}{5. Efficiency and Scalability}

CoMHR comprises 53.03M parameters in total, where the core hypergraph accounts for only 17.2\% ($\sim$9.1M), highlighting its parameter efficiency. On a single NVIDIA RTX 4090 GPU, our full relation reasoning stage takes approximately 28 ms per frame for standard groups (6-16 individuals). To address scalability in massive crowds (e.g., GigaCrowd), we employ a subgroup partitioning strategy (maximum $N=8$ per pass). This explicitly reduces the prohibitive $\mathcal{O}(N^{3})$ hypergraph complexity to strictly linear $\mathcal{O}(M)$ (where $M$ is the total number of individuals). Benchmarks validate this perfect linear scaling: processing each subgroup takes $\sim$28 ms, scaling exactly to $\sim$0.70 s for 200 individuals (25 subgroups). This guarantees a constant memory footprint regardless of crowd size, proving CoMHR's robust deployment potential without interaction-stage computational bottlenecks.

\subsection*{6. Limitations}
\addcontentsline{toc}{subsection}{6. Limitations}

Extensive stress tests demonstrate CoMHR's strong resistance to severe environmental degradation and upstream module failures. For instance, foreground truncation and sensor contamination increase the error by merely 2.41 mm and 1.18 mm, respectively. Similarly, extreme upstream corruptions, such as pelvis detection failure (altering error by +0.05 mm) and topology chaos injected into $T_z$ (+0.002 mm), are effectively isolated by our cross-modal architecture. 

However, CoMHR remains inherently dependent on the baseline foundation models (\eg, Depth Anything, OpenPose)~\cite{Yang2024DepthAU, Hidalgo2019OpenPoseW}. Its overall performance upper bound is intrinsically tied to their underlying dense feature extraction capabilities. Consequently, a primary limitation lies in simultaneous catastrophic collapse: in scenarios where all visual priors fail concurrently (\eg, extreme low light or 100\% occlusion), the model cannot recover the missing geometry and fails to reconstruct the human meshes. Furthermore, because CoMHR primarily focuses on spatial relational reasoning within a single frame, it lacks long-term temporal kinematic constraints. As a result, when individuals undergo prolonged periods of complete invisibility, the network cannot rely on historical motion trajectories to deduce their continuous states. We leave the integration of physics-based motion priors and generative temporal models to bridge these extreme observational gaps as a promising direction for future research.

\subsection*{7. Multimodal Contrastive Learning Pseudocode} 
\addcontentsline{toc}{subsection}{7. Multimodal Contrastive Learning Pseudocode} 

We present the pseudocode for the multimodal contrastive learning component of our CoMHR model to facilitate deeper understanding of our core training strategy. This pseudocode highlights the extraction of RGB, depth, and pose features, their relational encoding via PastEncoder, and the computation of intra- and cross-modal contrastive losses. While SMPL regression and mesh reconstruction are essential parts of the full model, they are handled separately and not included here. This focused pseudocode allows a clear illustration of how the model aligns and regularizes feature embeddings across multiple modalities to improve 3D reconstruction quality.

\begin{algorithm}[htbp]
\caption{Multimodal Contrastive Learning in CoMHR}
\label{alg:contrastive}
\begin{algorithmic}[1]
\Require RGB image $I$, depth map $D$, 3D keypoints $K$, validity mask $M$, temperature $\tau$
\Ensure Contrastive loss $\mathcal{L}_{con}$

\State Encode features:
\State \quad $f_{\text{rgb}} \gets \mathrm{RGBEncoder}(I)$
\State \quad $f_{\text{depth}} \gets \mathrm{DepthEncoder}(D)$
\State \quad $f_{\text{pose}} \gets \mathrm{PoseEncoder}(K, M)$

\State Build relational representations via PastEncoder:
\State \quad $r_{\text{rgb}} \gets \mathrm{PastEncoder}(f_{\text{rgb}})$
\State \quad $r_{\text{depth}} \gets \mathrm{PastEncoder}(f_{\text{depth}})$
\State \quad $r_{\text{pose}} \gets \mathrm{PastEncoder}(f_{\text{pose}})$

\State Normalize features: $\hat{r}_{*} \gets \mathrm{Normalize}(r_*)$  

\State Compute intra-modal contrastive loss for each modality:
\For{modality $m$ in $\{\text{rgb, depth, pose}\}$}
    \State Compute pairwise MPJPE matrix $D_{ij}$ between ground-truth poses
    \State Define positive set for sample $i$: $P_i = \{ j \mid D_{ij} < \text{thresh} \land i \neq j \}$
    \State Compute intra-modal loss (negatives masked out):
    \[
        \mathcal{L}_{\text{intra}}^m = - \frac{1}{N} \sum_{i} \frac{1}{|P_i|} \sum_{j \in P_i} \log \frac{\exp(\hat{r}_i \cdot \hat{r}_j / \tau)}{\sum_{k \in P_i} \exp(\hat{r}_i \cdot \hat{r}_k / \tau)}
    \]
\EndFor

\State Compute cross-modal contrastive loss:
\[
\mathcal{L}_{\text{cross}} = \mathrm{Mean}\Bigg( \mathrm{ReLU}\bigg( 
-\frac{1}{3} \sum_{m_1 < m_2} \hat{r}_{m_1} \cdot \hat{r}_{m_2} 
\bigg) \Bigg)
\]

\State Aggregate total contrastive loss:
\[
\mathcal{L}_{con} = \mathcal{L}_{\text{intra}}^{\text{rgb}} + \mathcal{L}_{\text{intra}}^{\text{depth}} + \mathcal{L}_{\text{intra}}^{\text{pose}} + \alpha \mathcal{L}_{\text{cross}}
\]

\State \Return $\mathcal{L}_{con}$
\end{algorithmic}
\end{algorithm}

\end{document}